\documentclass[]{article}
\usepackage[letterpaper]{geometry}
\usepackage{amta2022}
\usepackage{times}
\usepackage{url}
\usepackage{latexsym}
\usepackage{natbib}
\usepackage{layout}
\usepackage{etoolbox}

\AtBeginEnvironment{tabular}{\small}

\usepackage{amsmath,amsfonts,bm}

\def\eqref#1{equation~\ref{#1}}

\def\1{\bm{1}}

\DeclareMathAlphabet{\mathsfit}{\encodingdefault}{\sfdefault}{m}{sl}
\SetMathAlphabet{\mathsfit}{bold}{\encodingdefault}{\sfdefault}{bx}{n}

\usepackage{hyperref}
\usepackage{url}
\usepackage{booktabs}       
\usepackage{algorithm}
\usepackage{algorithmic}
\usepackage{amsfonts}

\usepackage[utf8]{inputenc} 
\usepackage[T1]{fontenc}    
\usepackage{hyperref}       
\usepackage{url}            
\usepackage{booktabs}       
\usepackage{amsfonts}       
\usepackage{nicefrac}       
\usepackage{microtype}      
\usepackage{xcolor}         

\usepackage{amsmath,bm}
\usepackage{stackrel}
\usepackage{bbm}
\usepackage{multirow}

\usepackage{pgfplots}
\pgfplotsset{compat=1.7}

\hypersetup{
  colorlinks, linkcolor=red, citecolor=cyan
}

\begin{document}

\title{\bf Quantized Wasserstein Procrustes Alignment of Word Embedding Spaces}

\newcommand*{\affmark}[1][*]{\textsuperscript{#1}}
\author{%
\parbox{\linewidth}{
\bf Prince Osei Aboagye\affmark[1]\thanks{~~work done while interning at Visa Research}\,\,\,, Yan Zheng\affmark[2], Chin-Chia Michael Yeh\affmark[2],  Junpeng Wang\affmark[2], Zhongfang Zhuang\affmark[2], Huiyuan Chen\affmark[2], Liang Wang\affmark[2], Wei Zhang\affmark[2], Jeff M. Phillips\affmark[1]} \\
\affmark[1]University of Utah, \affmark[2]Visa Research\\
\texttt{\affmark[1]\{prince\,,jeffp\}@cs.utah.edu}\\
\texttt{\affmark[2]\{yazheng\,,miyeh\,,junpenwa\,,zzhuang\,,hchen\,,liawang,wzhan\}@visa.com}
}

\maketitle
\pagestyle{empty}

\begin{abstract}

Optimal Transport (OT) provides a useful geometric framework to  estimate the permutation matrix under unsupervised cross-lingual word embedding  (CLWE) models that pose the alignment task as a Wasserstein-Procrustes problem. However, linear programming algorithms and approximate OT solvers via Sinkhorn for computing the permutation matrix come with a significant computational burden since they scale cubically and quadratically, respectively, in the input size. This makes it slow and infeasible to compute OT distances exactly for a larger input size, resulting in a poor approximation quality of the permutation matrix and subsequently a less robust learned transfer function or mapper. This paper proposes an unsupervised projection-based CLWE model called quantized Wasserstein Procrustes (qWP). qWP relies on a quantization step of both the source and target monolingual embedding space to estimate the permutation matrix given a cheap sampling procedure. This approach substantially improves the approximation quality of empirical OT solvers given fixed computational cost. We demonstrate that qWP achieves state-of-the-art results on the Bilingual lexicon Induction (BLI) task.

\end{abstract}

\section{Introduction}
\label{sec1}

In natural language processing (NLP), the problem of aligning monolingual embedding spaces to induce a shared cross-lingual vector space  has been shown not only to be useful in a variety of tasks such as bilingual lexicon induction (BLI) \citep{mikolov2013exploiting, barone2016towards, artetxe-etal-2017-learning, aboagye2022normalization}, machine translation \citep{artetxe2018iclr}, cross-lingual information retrieval \citep{vulic2015monolingual},  but it plays a crucial role in facilitating the cross-lingual transfer of language technologies from high resource languages to low resource languages.

Cross-lingual word embeddings (CLWEs) represent words from two or more languages in a shared cross-lingual vector space in which words with similar meanings obtain similar vectors regardless of their language. There has been a flurry of work dominated by the so-called \emph{projection-based} CLWE models \citep{mikolov2013exploiting, artetxe-etal-2016-learning, artetxe-etal-2017-learning, artetxe-etal-2018-robust, smith2017offline, Sebastian-Survey}, which aim to improve CLWE model performance significantly. \emph{Projection-based} CLWE models learn a transfer function or mapper between two independently trained monolingual word vector spaces with limited or no cross-lingual supervision.

Famous among \emph{projection-based} CLWE models are the unsupervised \emph{projection-based} CLWE models \citep{artetxe-etal-2017-learning, lample2018word, alvarez-melis2018gromov, pmlr-v89-grave19a}: they eliminate the initial seed bilingual lexicon and rely on the topological similarities between monolingual spaces, known as the isometry assumption, to extract seed bilingual lexicons. This makes them attractive since they require no cross-lingual supervision. One of the ways of framing unsupervised CLWE models is to pose the alignment task as a Wasserstein-Procrustes problem aiming to jointly estimate a permutation matrix and an orthogonal matrix \citep{pmlr-v89-grave19a, ramirez2020novel}. Most existing unsupervised CLWE models that solve the Wasserstein-Procrustes problem resort to Optimal Transport (OT) based methods to estimate the permutation matrix.

Optimal Transport (OT) \citep{monge1781memoire, kantorovich1942translocation} provides a natural geometric and probabilistic toolbox to compare probability distributions or measures. OT is concerned about determining an optimal transport plan for moving probability mass between two probability distributions with the cheapest cost. In theory, optimal transport is beautiful and well defined and has been well studied under continuous distribution. However, in practice or specifically in machine learning, we only have access to samples given an underlying distribution, so we turn to observe discrete distributions. This resonates with how empirical OT solvers have been built; they accept samples as inputs from input probability distributions or measures.

When the discrete distributions are composed of a large number of point cloud in higher dimensions, it becomes slow, impractical, and infeasible to compute OT distances exactly given the empirical OT solvers. A common scalable approach adopted by \citet{ pmlr-v89-grave19a} in their stochastic optimization framework to approximate the exact OT distance in order to extract the permutation matrix was to randomly draw $k$ monolingual embeddings from the source and target spaces, respectively. However, this approximation approach poses two main challenges:  
 
\paragraph{1) Sampling Efficiency} Does the OT distance computed between the $k$ sampled embeddings provide a useful or quality OT distance approximation of the true underlying distributions of the source and target spaces? Theoritical bounds and results have shown that the quality of this approximation has a convergence rate of $k^{-\frac{1}{d}}$ to the true OT distance, where $d$ is the ambient dimension \citep{dudley1969the, weed2019sharp}. Therefore, an effective approximation of the true OT distance requires large $k$ samples since we are constrained by the curse of dimensionality from the power ${-\frac{1}{d}}$. Thus, we need more samples to approximate the true OT distance in higher dimensions.

\paragraph{2) Computational Efficiency}  Empirical OT solvers such as linear programming algorithms \citep{Burkard} and approximate solvers via Sinkhorn \citep{NIPS2013_af21d0c9} for computing the permutation matrix have a computational cost of $\mathcal{O}\left(k^{3}\log k\right)$ and $\mathcal{O}\left(k^{2}\epsilon^{-2}\right)$, respectively, in the input size, $k$, and regularization term $\epsilon$ defined later in Equation \ref{eq:5}.  It becomes slow and infeasible in higher dimensions to compute OT distances exactly for a larger input size. We are therefore restricted by the maximum $k$ samples to draw for an effective approximation of the true OT distance. The constraint here is not the availability of data but computational cost.\\

Given these two challenges, \citet{beugnot2021improving} proposed two efficient OT estimators. The empirical OT solvers remain the same, either the linear programming solver or the entropic-regularized OT  via Sinkhorn. However, instead of drawing only $k$ samples as input to the OT solver, they rely on a cheap quantization step like $k$-means ++ \citep{arthur2006k} that is consistent with the computational complexity of the OT solver. Since sampling is cheap, they draw more than $k$ samples and then use  $k$-means++ to quantize the oversampled points from the source and target spaces, respectively, by partitioning them into $k$ clusters and then select the $k$ weighted anchor points as input to the  OT solver. This quantization step improves the approximation quality to the true OT distance.  Aside from the theoretical guarantees of the benefits of this quantization step, they showed that the new variant of the unregularized OT estimator yield an improvement in the convergence rate by $k^{-2\alpha}$ in the best case or $k^{-\alpha}$ in the worst case, which is on par with the computational complexity existing empirical OT estimators, where $\alpha=\frac{1}{d}$.

Inspired by the work of \citet{beugnot2021improving}, our paper proposes a new unsupervised CLWE model called quantized Wasserstein Procrustes (qWP). We follow the stochastic algorithm framework by \citet{pmlr-v89-grave19a} and the refinement procedure from \citet{lample2018word}.

\paragraph{Our contribution.} This work proposes a new unsupervised CLWE model: \textbf{quantized Wasserstein Procrustes (qWP)} that relies on a quantization step of the source and target distributions to estimate the alignment and linear transformation jointly. Firstly, we use the stochastic optimization framework in \citet{ pmlr-v89-grave19a}. However, instead of randomly drawing $k$ samples at each iteration, we use a quantization step to preprocess the source and target distributions to find the optimal $k$ point compression or summary needed to estimate the permutation matrix. It leads to a much-refined sample as opposed to a random sampling of the $k$ points. This approach substantially improves the approximation quality of the true OT distance and bias of empirical OT solvers given fixed computational cost \citep{beugnot2021improving}. The main idea behind qWP is to oversample the $k$ samples and then reduce them to $k$-weighted samples through quantization such as $k$-means++. After this, a linear program solver or regularized Sinkhorn algorithm can be used on the resulting quantized distribution. The translation pairs obtained from the permutation matrix are then used to learn the linear transformation. Finally, we use the refinement approach from \citet{lample2018word} to improve the orthogonal mapping. We demonstrate that qWP achieves state-of-the-art results on the BLI task. 

\section{Related Work} 
 
At the heart of Cross-lingual NLP are CLWE models. It has quickly evolved into a large subarea with a wide variety of approaches and perspectives, so we provide context by overviewing this work first.

Projection-based CLWE models can be categorized into \citep{Sebastian-Survey}: \textbf{1)} fully supervised projection-based CLWE models, \textbf{2)} weakly supervised projection-based CLWE models, and \textbf{3)} fully unsupervised projection-based CLWE models. The main idea governing all CLWE models is to independently train monolingual embeddings on large monolingual corpora in different languages or use pre-trained monolingual embeddings and then learn a transfer function to map them into a shared cross-lingual word vector space.

The first fully supervised projection-based CLWE model to learn a shared cross-lingual word vector space from monolingually-trained word embedding was proposed by \citet{mikolov2013exploiting}. They learned a linear transform from the source embedding space to the target language by minimizing the sum of squared Euclidean distance between the translation pairs of a seed dictionary based on the assumption that two embedding spaces exhibit similar geometric structures (i.e., approximately isomorphic). Their model requires word-level supervision from several thousand seed translation dictionaries (Dict). Subsequent works by \citet{xing2015normalized, artetxe-etal-2016-learning, reference_13} argued and proved that the quality of the learned CLWEs could be improved by modifying the objective function in \citet{mikolov2013exploiting}.


A more recent line of research has shown that the shared cross-lingual word vector space can be induced with weaker supervision from a small initial seed dictionary \citep{Vulic2016OnTR, glavas-etal-2019-properly, reference_11}. Weakly supervised projection-based CLWE models start with a small initial seed dictionary; however, the initial seed dictionary is iteratively expanded through a self-learning procedure. For example, Bootstrap Procrustes (PROC-B)~\citep{glavas-etal-2019-properly} is semi-supervised in that it starts with a small pairwise correspondence (of 500-1000 words), aligns those to infer a larger correspondence, and repeats applying Procrustes alignment. The quest to eliminate cross-lingual supervision has led to the development of fully unsupervised projection-based CLWE models. 

Fully unsupervised projection-based CLWE models use the topological similarities between monolingual embedding spaces to induce the shared cross-lingual vector space \citep{lample2018word, artetxe-etal-2018-robust, mohiuddin-joty-2019-revisiting}. The translation dictionaries are produced from scratch based on monolingual data only.

\section{Background}
\label{sec3}

In this section, we describe the mathematical formulation of supervised \emph{projection-based} CLWE models and unsupervised \emph{projection-based} CLWE models. We also defined what the 2-Wasserstein distance is and looked in detail at how the Wasserstein-Procrustes problem under the unsupervised CLWE model is solved in practice.

We define two monolingual embedding spaces as $X,Y\in\mathbb{R}^{n\times d}$, where $n$ is the number of words, and $d$ is the dimension of the monolingual word embeddings.

\paragraph{Supervised \emph{Projection-Based} CLWE Models} require word-level supervision from seed translation dictionaries such that word $x_{i}$ in $X$ is the translation of word $y_{i}$ in $Y$. The linear transformation, $W^{\ast}$, from the source monolingual embedding space to the target monolingual embedding space is learned by solving the least square problem \citep{mikolov2013exploiting}:

\begin{equation}
W^{\ast}=\underset{W\in\mathbb{R}^{d\times d}}{\arg\min}\left\Vert XW-Y\right\Vert _{F}^{2}\label{eq:1}
\end{equation}

\cite{xing2015normalized}, modified the objective function in Eq. (\ref{eq:1}) to improve the quality of the learned CLWEs  by unit length normalizing the word embeddings and imposing an orthogonality constraint on the linear transformation ($W$) during training:

\begin{equation}
W^{\ast}=\underset{W\in\mathcal{O}_{d}}{\arg\min}\left\Vert XW-Y\right\Vert _{F}^{2},\label{eq:2}
\end{equation}

where $\mathcal{O}_{d}$ is the set of orthogonal matrices. The orthogonality constraint preserves the original monolingual embedding space's similarities and geometric structure.
These assumptions and constraints imposed on the linear transform make the problem of learning a transfer function an orthogonal \textbf{Procrustes} problem (Eq. \ref{eq:2}), which has a closed-form solution: $W^{\ast}=UV^{\top},$ where $U\varSigma V^{\top}$ is the singular
value decomposition of $X^{\top}Y$ \citep{schonemann1966generalized}.

\paragraph{2-Wasserstein distance}  is a distance function used to compute the OT-distance given two set of points $X$ and $Y$:

\begin{equation}
W_{2}^{2}\left(X,Y\right)=\underset{P\in\mathcal{P}_{n}}{\min}\stackrel[i,j=1]{n}{\sum}\left\Vert x_{i}-y_{j}\right\Vert _{2}^{2}P_{ij}\label{eq:4*}
\end{equation}
where $\mathcal{P}_{n}$ is the set of permutation matrices, $\mathcal{P}_{n}=\left\{ P\in\left\{ 0,1\right\} ^{n\times n},\,P1_{n}=1_{n},\,P^{\top}1_{n}=1_{n}\right\} $.

\paragraph{Unsupervised \emph{Projection-Based} CLWE Models}  

Without any initial seed bilingual lexicon some unsupervised CLWE models solves the \textbf{Wasserstein-Procrustes} problem (Eq. \ref{eq:3})  to jointly estimate the permutation matrix or alignment ($P$) and  linear transformation ($W$)  \citep{pmlr-v89-grave19a, ramirez2020novel}: 

\begin{equation}
W^{\ast},P^{\ast}=\underset{W\in\mathcal{O}_{d},P\in\mathcal{P}_{n}}{\arg\min}\left\Vert XW-PY\right\Vert _{F}^{2}\label{eq:3}
\end{equation}
The permutation matrix $P^{*}$  provides a one-to-one mapping or correspondence between the source and target samples.

Under unsupervised CLWE models that solve the Wasserstein-Procrustes problem, we aim to estimate the two unknown variables $W$ and $P$. One way to solve Eq. (\ref{eq:3}) is by alternating the
minimization of  $W$ and $P$. Given $P$, we use the translation pairs obtained between the source and target spaces to learn the linear transformation, $W^{*}$ from Eq. (\ref{eq:2}). Similarly, given the linear transformation $W^{*}$, Eq. (\ref{eq:3}) is equivalent to minimizing the 2-Wasserstein distance between $XW$ and $Y$ to solve for the permutation matrix, $P$:

%
%
%
%

\begin{equation}
W_{2}^{2}\left(XW,Y\right)=\underset{P\in\mathcal{P}_{n}}{\min}\stackrel[i,j=1]{n}{\sum}\left\Vert x_{i}W-y_{j}\right\Vert _{2}^{2}P_{ij}\label{eq:4}
\end{equation}

Equation (\ref{eq:4}) is the standard OT problem, and it can be solved using a linear programming solver, which has a computational cost of $\mathcal{O}\left(n^{3}\log n\right)$. For a large $n$, a linear programming solver is impractical. Another variant and approximation of the optimal transport problem were proposed by \citep{NIPS2013_af21d0c9}. This variant adds an entropic regularization term leading to the Sinkhorn algorithm with a computational cost of $\mathcal{O}\left(n^{2}\epsilon^{-2}\right)$:

\begin{equation}
W_{2}^{2}\left(XW,Y\right)=\underset{P\in\mathcal{P}_{n}}{\min}\stackrel[i,j=1]{n}{\sum}\left\Vert x_{i}W-y_{j}\right\Vert _{2}^{2}P_{ij} +\epsilon\stackrel[i,j=1]{n}{\sum}\log P_{ij} \label{eq:5*}
\end{equation}

\cite{pmlr-v89-grave19a} proposed a stochastic optimization scheme to jointly estimate $W$ and $P$ by randomly sampling $\hat{X},\hat{Y}\in\mathbb{R}^{k\times d}$ from $X$ and $Y$, where $k<n.$ Due to how slow and infeasible a linear programming solver for a larger input size can be,  \cite{pmlr-v89-grave19a} used the Sinkhorn algorithm to compute the permutation matrix, $P$  by minimizing:

\begin{equation}
W_{2}^{2}\left(\hat{X}W,\hat{Y}\right)=\underset{P\in\mathcal{P}_{k}}{\min}\stackrel[i,j=1]{k}{\sum}\left\Vert x_{i}W-y_{j}\right\Vert _{2}^{2}P_{ij} +\epsilon\stackrel[i,j=1]{k}{\sum}\log P_{ij} \label{eq:5}
\end{equation}

\section{Proposed Method}
 
This section introduces our new unsupervised CLWE model: quantized Wasserstein Procrustes (qWP). We use the previous stochastic algorithm framework and refinement procedure from \cite{pmlr-v89-grave19a} and \citet{lample2018word} respectively in our model, but we rely on a quantization step to estimate the permutation matrix.

\subsection{quantized Wasserstein Procrustes (qWP)}

We consider two languages with vocabularies $V_{x}$ and $V_{y}$, represented by word embeddings $\text{X}=\left\{ x_{i}\right\} _{i=1}^{n}$,$\,\text{Y}=\left\{ y_{i}\right\} _{i=1}^{n}$, respectively. We assume  two empirical distributions over the embedding spaces, $X$ and $Y$: $\mu=\stackrel[i=1]{n}{\sum}p_{i}\delta_{x^{\left(i\right)}}$ and $\nu=\stackrel[j=1]{n}{\sum}q_{j}\delta_{y^{\left(j\right)}}$, where $p_{i}$ and $q_{i}$ are the probability weights associated with each word vector, $\delta_{x}$ and $\delta_{y}$ is the Dirac function supported on point $x$ and $y$ respectively.

The main crux of our proposed unsupervised CLWE model: quantized Wasserstein Procrustes (qWP) is that we rely on a quantization step like $k$-means++~\citep{arthur2006k} instead of random sampling to estimate the permutation matrix and then use gradient descent and Procrustes to extract the orthogonal matrix. We take Eq. (\ref{eq:3}) as our loss function. However, Eq. (\ref{eq:3}) is not jointly convex in $W$ and $P$, but as we saw in Section \ref{sec3} we can fix one variable and then solve for the other variable. Alternating the minimization in each variable $W$ and $P$ is therefore employed to find a solution \citep{alaux2018unsupervised, pmlr-v89-grave19a}.

First, we have to induce the translation dictionary by solving for the permutation matrix, $P^{\ast}$  in Eq. (\ref{eq:4}) and then find the orthogonal projection matrix from Eq. (\ref{eq:2}). Naively doing an alternating full minimization in each variable $W$ and $P$ of Eq. (\ref{eq:3}) does not scale, and even on smaller problems, empirical results show that it quickly converges to a bad local minima \citep{zhang-etal-2017-earth}. A scalable stochastic approach adopted by \citet{pmlr-v89-grave19a} was to instead, at each iteration, $t$, randomly sample a minibatch $\ensuremath{\text{X}_{k}=\left\{ x_{i}\right\} _{i=1}^{k}},\ensuremath{\,\text{and Y}_{k}=\left\{ y_{i}\right\} _{i=1}^{k}}$  of size $k$ from $X$ and $Y$. The optimal coupling or permutation matrix, $P^{\ast}$, was then computed from Eq. (\ref{eq:5}) using the Sinkhorn algorithm. The translation pairs obtained from $P^{\ast}$ between the source and target spaces are then used to learn the orthogonal matrix, $W^{*}$, that maps the source to the target spaces from Eq. (\ref{eq:2}) by using Procrustes and gradient descent to update $W$. The procedure for updating $W$ is detailed in \citet{pmlr-v89-grave19a}.

\begin{figure}
    \includegraphics[width=\linewidth]{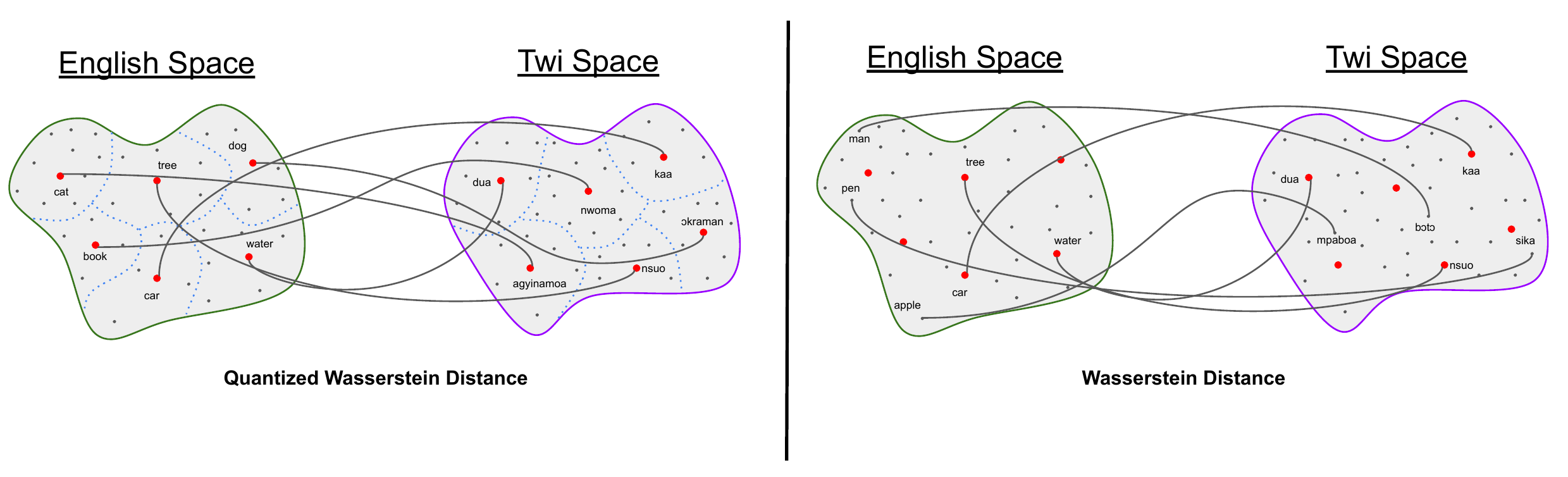}
    \vspace{-.3in}
    \caption{Illustration on toy $2d$ data showing the potential advantage of Quantized Wasserstein Distance (qWD) over Wasserstein Distance (WD). We want to align or translate words in the English Space to words in the Twi Space without knowing aforehand the translation pairs or the linear transformation. Twi is a language spoken in Ghana, West Africa. First, we must induce the translation pairs by estimating the permutation matrix, $P$, either through qWD or WD. Each dot represents a word in that space; specifically, the red points are the $k$ centers from $k$-means++. The edge connecting two red points means the two words are accurate translation pairs, whereas the edge between two black points is the wrong translation pair. Here we want to induce six translation pairs through $P$.} 
    \label{fig:SN-examp}
\end{figure}

 The stochastic optimization scheme adopted by \citet{pmlr-v89-grave19a} to make the alternating minimization process scale and achieve a better convergence to a good local minimum when computing the permutation matrix suffers from the sampling efficiency and computational efficiency challenges discussed in Section \ref{sec1}.

To address these two challenges following \citet{beugnot2021improving}, we will quantize the source and target word embedding space by finding the optimal $k$ point compression or summary as input to the 2-Wasserstein distance \citep{Pollard, Canas} through the use of $k$-means++. The resulting convergence rate of $k^{-2\alpha}$ in the best case or $k^{-\alpha}$ in the worst case from using this quantization step makes the OT solver yields a better approximation quality of the permutation matrix and subsequently a more robust learned transfer function, where $\alpha=\frac{1}{d}$.

\begin{algorithm}[ht]
    \caption{Quantized Wasserstein Procrustes}
        \label{Alg1}

    \begin{algorithmic}[1]
    
\REQUIRE \quad \\
 Word embedding matrix, $X, Y \in \mathbb{R}^{n \times d}$ of the source language and target language  respectively, entropy regularization coefficient $\epsilon$, number of anchor point $k$\\
 \ENSURE \quad \\ Orthogonal matrix, $\text{W}$
 
\FOR {$e=1,\ldots,E$} 
    \FOR {$t=1,\ldots,T$} 
  \STATE $P\leftarrow qW\left(X,Y,\epsilon,k\right)$
  \STATE $W\leftarrow$Update $W$ by gradient descent and Procrutes
    \ENDFOR
\ENDFOR  
\RETURN    $W$
\end{algorithmic}
\end{algorithm}

\subsubsection{New Alignment Algorithm}


The goal of our proposed new CLWE algorithm is to quantize the source and target embedding spaces $X$ and $Y$ to be aligned to obtain a much-refined coreset \footnote{A coreset is a summary or an approximation of the shape of a larger point cloud with a smaller point cloud.} that is less noisy compared to just randomly sampling from $X$ and $Y$. Our proposed new method is summarized in Algorithms \ref{Alg1} and \ref{Alg2}. For each iteration $t$ (Algorithm \ref{Alg1}), we compute the permutation matrix $P^{\ast}$ from Algorithm \ref{Alg2}. The main idea of Algorithm \ref{Alg2} is to draw more than $k$ samples using the coreset size $m>k$
and then reduce them to $k$-weighted samples through quantization such as $k$-means++. Here the computational cost of $k$-means++ is $\mathcal{O}\left(mk\right)$. To satisfy the computational complexity of the OT solver, we must ensure that the quantization step used to preprocess the source and target space takes $\mathcal{O}\left(k^{3}\log k\right)$ time.  In view of this, we set $m=k^{2}\log k$ so that we are consistent with the computational complexity $\mathcal{O}\left(k^{3}\log k\right)$ of the OT solver. We then sample $\text{X}_{m}=\left(x_{1},\ldots x_{m}\right)$ i.i.d from $X$ and $\text{Y}_{m}=\left(y_{1},\ldots y_{m}\right)$ i.i.d from $Y$. Using $k$-means++ we find the $k$ weighted centers. Following each Voronoi cell, we weight each center proportionally to the number of samples to obtain the weights $a$ and $b$. We then can use either the linear program solver or the regularized Sinkhorn algorithm \citep{NIPS2013_af21d0c9} to estimate the permutation matrix, $P$, between the two quantized point clouds. In our case, we used the entropic-regularized OT solver  via Sinkhorn, which we call A{\scriptsize{}PPROX}OT$\left(C,a,b,\epsilon\right)$.

\begin{algorithm}[h]
    \caption{Quantized $2\text{-Wasserstein }\text{Distance}\, (qW\left(X,Y,\epsilon,k\right))$}
    \begin{algorithmic}[1]
\REQUIRE \quad \\
$\text{X}=\left\{ x_{i}\right\} _{i=1}^{n}$,$\,\text{Y}=\left\{ y_{i}\right\} _{i=1}^{n},\,\text{entropy regularization coefficient } \epsilon,\,\text{number of anchor points }k$\\ 
\ENSURE \quad \\
Permutation Matrix, $P$\\
        \STATE  Sample $m$ points:
        \STATE $\,\,\,\,\,$Set $m=k^{2}\log k$
         \STATE $\,\,\,\,\,$Sample $\text{X}_{m}=\left(x_{1},\ldots x_{m}\right)$ i.i.d from $X$  and $\text{Y}_{m}=\left(y_{1},\ldots y_{m}\right)$  i.i.d from $Y$
\STATE Subsample $k$ anchor points:
\STATE  $\,\,\,\,\,$ Compute $\left(c_{1},\ldots c_{k}\right)$ with $k-$means++
\STATE $\,\,\,\,\,$ Compute $\left(d_{1},\ldots d_{k}\right)$ with $k-$means++
\STATE Compute weights:
\STATE $\,\,\,\,\,$ Set $a_{i}=\stackrel[j=1]{n}{\sum}\boldsymbol{1}_{i=\underset{l}{\arg\min}\left\Vert x_{j}-c_{l}\right\Vert _{2}^{2}}\,\forall i\in\left\{ 1,\ldots,k\right\} $
\STATE $\,\,\,\,\,$ Set $b_{i}=\stackrel[j=1]{n}{\sum}\boldsymbol{1}_{i=\underset{l}{\arg\min}\left\Vert x_{j}-d_{l}\right\Vert _{2}^{2}}\,\forall i\in\left\{ 1,\ldots,k\right\} $
\STATE Cost matrix:
\STATE $\,\,\,\,\,$ Set $C_{ij}=\left\Vert c_{i}-d_{j}\right\Vert _{2}^{2}\,\forall i,j\in\left\{ 1,\ldots,k\right\}$  
\STATE Regularized transport solver: \\
\RETURN    $P\leftarrow$ A{\scriptsize{}PPROX}OT$\left(C,a,b,\epsilon\right)$

    \end{algorithmic}
    \label{Alg2}
\end{algorithm}

See the example in Figure \ref{fig:SN-examp} where the translation pairs obtained under qWD yield perfect matches compared to WD, which gave some wrong translation pairs. Under qWD we use $k$-means++ to quantize the English and Twi Space to select the $k$ weighted centers as input to the OT solver instead of randomly drawing $k$ points under WD, which could be noisy.

As a quick review of $k$-means++~\citep{arthur2006k}, it initializes a set of cluster centers for the $k$-means objective. Each step iteratively increases the set of cluster centers by choosing a new center from the dataset proportional to the squared distance to the closest already chosen center. In one variant we explore, we run one step of the standard Lloyd’s algorithm after initializing, moving each center found to the average of data points closest to it.



\section{Experimental Analysis}

We provide an evaluation of our proposed methods using English (EN) and five languages embeddings pre-trained on Wikipedia \citep{bojanowski-etal-2017-enriching}:  Spanish (ES), French (FR), German (DE), Russian (RU), and Italian (IT). We use the 300-dimensional fastText~\citep{bojanowski-etal-2017-enriching}  embeddings, and all vocabularies are trimmed to the 200K most frequent words.


\paragraph{Alignment evaluation tasks: BLI} We evaluate and compare our proposed CLWE method mainly on the Bilingual Lexicon Induction (BLI) task, a word translation task. BLI is more direct and has become the de facto evaluation task for CLWE models. For words in the source language, this task retrieves the nearest neighbors in the target language after alignment to check if it contains the translation. We report two different translation accuracies: precision at 1 (P@1) and mean average precision (MAP) \citep{glavas-etal-2019-properly} translation accuracy, which is equivalent to the mean reciprocal rank (MRR) of the translation.

\paragraph{Implementation Details} The monolingual word embeddings are unit length normalized and centered before entering the model. The first 2.5k words are used to determine $Q_{0}$ given $P^{*}$ obtained from the Frank-Wolfe algorithm \citep{FrankWolfe}. We trained qWp on the first 20k most frequent words and evaluated them on separate 1.5k source test queries. We used the MUSE publicly available
translation dictionary \citep{lample2018word}. We used the regularized Sinkhorn algorithm \citep{NIPS2013_af21d0c9} and always set the entropy regularization term ($\epsilon$) to $\epsilon = 0.05$.

We use the \emph{Refinement} approach from \citep{lample2018word} and run it for five epochs. This approach iteratively improves the orthogonal mapping $Q$.   After learning $Q^{*}$ from Eq. (\ref{eq:3}), we build another (slightly larger) dictionary of translation pairs by translating each word to its nearest neighbor under the transformation $Q$. The newly learned dictionary of translation pairs is then used to learn a new mapping $Q$ from Eq. (\ref{eq:2}), and then we repeat the process, each time building an incrementally larger dictionary.

We consider both balanced and unbalanced OT. The unbalanced OT does not require strict mass preservation \citep{chizat2018interpolating}, contrary to the standard or balanced OT problem, Eq. (\ref{eq:4}). Under the unbalanced OT, Eq. (\ref{eq:4}) is relaxed by adding two KL-divergence terms to ensure a more relaxed mass preservation. This helps to solve the polysemy problem.


\begin{table}[ht]
  \caption{Bilingual lexicon Induction (BLI) task, (MAP) - Without Refinement} 
  \label{tab:BLI-MAP-woR}
\centering

  \begin{tabular}{llccccc}
  \toprule
&  \multicolumn{4}{r}{Coreset Size} \\
\cmidrule(l){3-7}
Trans. Pairs     & Sampling   & 200 &  500  & 1000 &  2000 & 3000      \\
    \midrule
    
    \multirow{2}*{EN-ES}  & Random &36.40   &47.22    &48.90   &49.66   &50.08             \\   
                              & KMeans ++     &45.21   &48.69   &49.64   &49.71    &50.09              \\  
 
    \multirow{2}*{ES-EN}  & Random &43.74   &50.36   &52.04  &53.84   &54.67         \\   
                              & KMeans ++     &47.24   &52.21    &52.55   &54.10   &54.90              \\

    \midrule

    \multirow{2}*{EN-FR}  & Random &37.22   &47.94    &49.31   &50.59   &50.88            \\   
                              & KMeans ++     &46.54  &49.45    &50.12   &50.54    &51.07         \\  

    \multirow{2}*{FR-EN}  & Random &38.87  &53.36    &54.91   &55.47    &55.85             \\   
                              & KMeans ++     &52.67   &54.43    &55.54   &56.11    &56.70        \\  
                              
    \midrule

    \multirow{2}*{EN-DE }  & Random &27.18   &36.10    &38.00   &38.49    &39.29      \\   
                              & KMeans++     &32.80   &37.44    &38.58  &38.77   &39.72        \\  
                              
    \multirow{2}*{DE-EN}  & Random &30.97   &41.26    &40.44  &41.87    &41.78          \\   
                              & KMeans ++     &39.03   &41.90    &40.21  &43.93    &42.42            \\  
 
    \midrule  
    
    \multirow{2}*{EN-RU }  & Random &18.68   &27.91    &30.91  &31.75    &32.43               \\   
                              & KMeans ++     &26.97   &27.12    &29.90   &32.25    &31.41              \\  
                              
    \multirow{2}*{RU-EN}  & Random &27.26   &39.93    &41.50   &43.56    &43.81              \\   
                              & KMeans ++     &16.13  &37.07    &42.69   &42.82    &44.54            \\  
 
     \midrule

    \multirow{2}*{EN-IT}  & Random &34.04   &46.10   &47.99   &49.28    & 50.79           \\   
                              & KMeans ++     &44.83   &47.31    &49.00   &50.29    &51.12         \\  
                              
    \multirow{2}*{IT-EN}  & Random &38.50  &52.44   &52.92   &54.70   &57.04         \\   
                              & KMeans ++     &47.80   &51.77    &54.60  & 57.03   & 57.49        \\

     \midrule  
         \midrule  
    
    \multirow{2}*{Avg}  & Random & 33.28  &  44.26  &45.69    &46.92    & 47.66          \\ 
                              & KMeans ++     & \textbf{39.92}   & \textbf{44.74}    &\textbf{46.28}    &\textbf{47.55}   &\textbf{47.94}        \\

    \bottomrule

\end{tabular}\\
\end{table}

\paragraph{Baselines: BLI} We evaluated and compared the published result of qWP to several supervised and unsupervised CLWE models on the BLI task. The baselines include Procrustes (PROC)~\citep{artetxe-etal-2016-learning}, Ranking-Based Optimization (RCSLS)~\citep{joulin-etal-2018-loss}, Gromov Wasserstein (GW) \citep{alvarez-melis2018gromov}, Adversarial Training (Adv + Refine) \citep{lample2018word} and the density matching method (Dema + Refine) \citep{Wang2019WassersteinFisherRaoDD}. We used the baseline results

\paragraph{Main Results}  Tables \ref{tab:BLI-MAP-woR}, \ref{tab:BLI-P1-woR}  and \ref{tab:BLI-MAP-wR} summarize the effect of the coreset size within the qWP algorithm. We proceed with four experiments. In tables \ref{tab:BLI-MAP-woR} and \ref{tab:BLI-MAP-wR} we report the mean average precision (MAP) \citep{glavas-etal-2019-properly} translation accuracy, which is equivalent to the mean reciprocal rank (MRR) of the translation, whereas, Tables  \ref{tab:BLI-P1-woR} and \ref{tab:BLI-comp}, the translation accuracy reported is the precision at 1 (P@1).

\begin{table}[H]
  \caption{Bilingual lexicon Induction (BLI) task, (P@1) Without Refinement} 
  \label{tab:BLI-P1-woR}
\centering

  \begin{tabular}{llcccc}
  \toprule
&  \multicolumn{4}{r}{Coreset Size} \\
\cmidrule(l){3-6}
Translation Pairs     & Sampling   & 500 &  1000  & 2000 &  3000     \\
    \midrule
    
    \multirow{2}*{EN-ES}  & Random &73.53  &75.20  &76.73  &80.40       \\   
                              & KMeans ++     &77.80  &79.47  &78.20  &81.53         \\  
 
    \midrule

    \multirow{2}*{EN-FR}  & Random &77.07  &79.40  &80.00  & 81.00       \\   
                              & KMeans ++     &78.27  &79.60  &80.20  &81.13        \\  
 
    \midrule

    \multirow{2}*{EN-DE }  & Random &62.73  &67.60  &70.40  &70.60      \\   
                              & KMeans++     &65.60  &68.87  &71.40  &71.40         \\  
 
    \midrule  
    
    \multirow{2}*{EN-RU }  & Random &33.13  &35.53  & 35.47 &36.87           \\   
                              & KMeans ++     &34.53  &36.07 &36.53  & 36.60          \\  
 
    \midrule

    \multirow{2}*{EN-IT}  & Random &70.67  &72.47  &75.13  &75.73        \\   
                              & KMeans ++     &73.20  &74.87  &76.73  & 76.93         \\  
 
    \midrule  
        \midrule 
    
    \multirow{2}*{Avg}  & Random &63.43   &66.04   &67.55   & 68.92        \\   
                              & KMeans ++   &\textbf{65.88}   &\textbf{67.78}   &\textbf{68.61}   &\textbf{69.52}           \\  
    
    \bottomrule

\end{tabular}\\
\end{table}

The first experiments in Table \ref{tab:BLI-MAP-woR} show the MRR scores without refinement, and the following Table \ref{tab:BLI-MAP-wR} shows the same MRR scores with refinement. In each table, we increase the coreset size from 200 to 3000, and this is either chosen as in prior work as a random sample or in our proposed approach via $k$-means++. As expected, on all language pairs, the performance increases as the coreset size increases. Also, notice that the improvement by increasing the coreset size plateaus and is not as significant from 2000 to 3000, indicating that probably 2000 coreset points are usually sufficient.

We also observe that in almost all cases, the performance is improved when using the $k$-means++ coreset instead of the random sample coreset.  The few exceptions are mostly in the comparison with Russian 
(RU) with refinement, but this gap narrows as the coreset size increases.  Notably, by coreset size of 2000, the $k$-means++ coresets have a clear advantage with an average improvement of from 46.92 to 47.55 without refinement and from 53.05 to 53.76 with refinement. This follows the general trend of better scores when the refinement phase is used.

Table \ref{tab:BLI-P1-woR} shows a similar experiment on the BLI tasks but reports the precision at 1 (P@1) score. The results show a strong average improvement while using $k$-means++, with the exception being EN-RU with a small advantage of random sampling at 3000 coreset size; however, with MAP, the results for $k$-means++ are already basically as good with 2000 points.




\begin{table}[H]
  \caption{Bilingual lexicon Induction (BLI) task, (MAP) With Refinement} 
  \label{tab:BLI-MAP-wR}
\centering

  \begin{tabular}{llccccc}
  \toprule
&  \multicolumn{4}{r}{Coreset Size} \\
\cmidrule(l){3-7}
Trans. Pairs     & Sampling   & 200 &  500  & 1000 &  2000 & 3000       \\
    \midrule
    
    \multirow{2}*{EN-ES}  & Random & 54.45  & 54.35   & 54.54  & 54.56   & 54.61         \\   
                              & KMeans ++     & 54.41  &54.48   & 54.55  & 54.67   & 54.72            \\  
 
    \multirow{2}*{ES-EN}  & Random & 60.96  & 58.24   & 58.56  & 58.88   & 59.69           \\   
                              & KMeans ++     & 58.01  & 58.26   & 59.22  & 59.11   &59.55            \\

    \midrule

    \multirow{2}*{EN-FR}  & Random & 54.93  & 55.26   & 55.31  & 55.31   & 55.24          \\   
                              & KMeans ++     & 55.05  & 55.41   & 55.44  & 55.38   & 55.30        \\  

    \multirow{2}*{FR-EN}  & Random & 56.00  & 61.36   & 61.44  & 61.46   & 61.51            \\   
                              & KMeans ++     & 61.81  & 61.68   & 61.54  & 61.60   & 61.64       \\  
                              
    \midrule

    \multirow{2}*{EN-DE }  & Random & 43.42  & 43.28   &43.42  & 43.46   & 43.37       \\   
                              & KMeans++     & 43.12  & 43.32   &43.56  & 43.59   & 43.52      \\  
                              
    \multirow{2}*{DE-EN}  & Random & 48.45  & 48.70   & 45.74  & 46.03   &46.72          \\   
                              & KMeans ++     & 45.91  & 49.05   & 46.69  & 48.78   & 48.54          \\  
 
    \midrule  
    
    \multirow{2}*{EN-RU }  & Random & 40.34  & 41.56   &42.92  & 42.50   & 42.76              \\   
                              & KMeans ++     & 41.57  & 40.08   & 41.41  & 43.07   & 41.39             \\  
                              
    \multirow{2}*{RU-EN}  & Random & 48.01  & 49.28   & 48.64  & 50.09   & 50.48       \\   
                              & KMeans ++     & 38.69  & 46.24   & 50.16  & 49.05   & 50.43           \\  
 
    \midrule

    \multirow{2}*{EN-IT}  & Random & 55.93  & 56.82   & 57.36  & 57.48   &  57.54          \\   
                              & KMeans ++     & 56.23  & 56.55   & 57.32  & 57.75   & 57.41         \\  
                              
    \multirow{2}*{IT-EN}  & Random & 59.71  & 61.44   & 60.22  & 60.70   & 65.10     \\   
                              & KMeans ++     & 60.13  & 59.55   & 60.61  &  64.62  &  64.71         \\

    \midrule  
        \midrule 
    
    \multirow{2}*{Avg}  & Random & \textbf{52.22}   &\textbf{53.02}    &52.82    &53.05    &53.70           \\   
                             & KMeans ++ &51.49   &52.46    &\textbf{53.05}    &\textbf{53.76}    &\textbf{53.72}           \\  
 
    \bottomrule

\end{tabular}\\
\end{table}

The final experiment in Table \ref{tab:BLI-comp} shows the results of our proposed methods against state-of-the-art techniques. We used a fixed coreset size of 2000. Each entry shows the P@1 scores on the BLI task. The first two lines show PROC and RCSLS, which are supervised methods, so they know the alignment between 5000 pairs of works across embeddings and use this knowledge to determine the alignment. Notice our techniques (which are unsupervised) improve upon the standard Procrustes alignment (PROC) and are almost competitive with the RCSLS method, which optimizes for the BLI task specifically.

Our method also outperforms Gromov-Wasserstein (GW) alignment, as well as Adv + Refine, Dema + Refine, and a random sample coreset when using refinement.

In this table, we also show experiments with two other enhancements. The first is to improve the cluster centers and the quantization found with $k$-means++ with a run of Lloyd’s algorithm (the standard $k$-means optimization procedure) for 1 step. This moves the quantization point to the center of the points it represents, making it more representative on average. This provides a small improvement. The second extension is to use unbalanced optimal transport instead of balanced OT. Surprisingly, this offers no advantage on average.

\begin{table}[H]
 \setlength{\tabcolsep}{5pt}
\caption{Bilingual lexicon Induction (BLI) task, Comparison with other Methods}
\label{tab:BLI-comp}
\begin{center}
\begin{tabular}{lccccccc} 
    \toprule
Method &  & EN-ES & EN-FR & EN-DE & EN-RU & EN-IT & Avg\\
    \midrule
 & Dict & $\rightarrow\,\,\,\leftarrow$ & $\rightarrow\,\,\,\leftarrow$ & $\rightarrow\,\,\,\leftarrow$ & $\rightarrow\,\,\,\leftarrow$ & $\rightarrow\,\,\,\leftarrow$ & \\   
    \midrule
PROC & 5K  & $81.9\,\,\,83.4$ & $82.1\,\,\,82.4$ & $74.2\,\,\,72.7$ & $51.7\,\,\,63.7$ & $77.4\,\,\,77.9$ & 74.7\\   
RCSLS & 5K &$ 84.1\,\,\,86.3$ & $83.3\,\,\,84.1$ & $79.1\,\,\,76.3$ & $57.9\,\,\,67.2$ & &  77.3\\   
    \midrule            
GW & None & $81.7\,\,\,80.4$ & $81.3\,\,\,78.9$ & $71.9\,\,\,78.2$ & $45.1\,\,\,43.7$ & $78.9\,\,\,75.2$ & 71.5\\     
Adv + Refine & None & $81.7\,\,\,83.3$ & $82.3\,\,\,82.1$ & $74.0\,\,\,72.2$ & $44.0\,\,\,59.1$ & $77.9\,\,\,77.5$ & 73.4\\
Dema + Refine & None & $82.8\,\,\,84.9$ & $82.6\,\,\,82.4$ & $75.3\,\,\,74.9$ & $46.9\,\,\,62.4$ &  & 74.0\\
\midrule 
Random  \\
WP + Refine & None & $82.8\,\,\,84.1$ & $82.6\,\,\,82.9$ & $75.4\,\,\,73.3$ & $43.7\,\,\,59.1$ &  & 73.0\\


\midrule 
\textbf{Unbalanced OT}\\
\midrule 
(Ours) KMeans++\\
qWP + Refine& None & $83.9\,\,\,84.5$ & $83.6\,\,\,83.1$ & $77.0\,\,\,74.9$ & $48.0\,\,\,60.1$ & $80.5\,\,\,80.7$ &\textbf{75.6} \\   

\midrule 
(Ours) LloydRefine\\

qWP + Refine& None & $83.8\,\,\,84.9$ & $84.3\,\,\,83.4$ & $77.0\,\,\,75.2$ & $48.2\,\,\,61.3$ &  $80.5\,\,\,80.9$ & \textbf{75.9}\\   

\midrule 
\textbf{Balanced OT}\\
\midrule 
(Ours) KMeans++\\
qWP + Refine& None & $83.5\,\,\,84.3$ & $84.0\,\,\,83.1$ & $76.9\,\,\,74.9$ & $46.6\,\,\,59.8$ & $80.6\,\,\,80.3$ & \textbf{75.4}\\   

\midrule 
(Ours) LloydRefine\\

qWP + Refine& None & $83.6\,\,\,84.4$ & $84.0\,\,\,83.1$ & $77.1\,\,\,74.8$ & $47.3\,\,\, 60.4$ & $80.1\,\,\,80.4$ & \textbf{75.5}\\   

\bottomrule    
    
  \end{tabular}
\end{center}
\end{table}

\section{Conclusion}
This paper presents an approach to aligning embeddings in high-dimensional space. While the overall problem is non-convex and computationally expensive, we present an efficient stochastic algorithm to solve the problem based on a refined sample set. This paper focuses on the matching procedure of the BLI task. Our key insight is that our quantization algorithm can outperform the current state-of-art unsupervised algorithm on both balanced and unbalanced settings of the loss function.


\newpage

\bibliography{amta2022}
\bibliographystyle{amta2022}



\end{document}